\documentclass[letterpaper, 10 pt, conference]{ieeeconf}  

\IEEEoverridecommandlockouts                              

\usepackage{graphics} 
\usepackage{epsfig} 
\usepackage{amssymb}  

\title{\LARGE \bf
Weakly Supervised Point Clouds Transformer for 3D Object Detection
}

\author{Zuojin Tang$^{1,2}$, Bo Sun$^{2*}$, Tongwei Ma$^{3}$, Daosheng Li$^{3}$, Zhenhui Xu$^{3}$
\thanks{$^{1}$College of Software Engineering, Southeast University}%
\thanks{$^{2}$Quanzhou Institute of Equipment Manufacturing, Haixi Institutes, Chinese Academy of Sciences}%
\thanks{$^{*}$Correspondence: sunbo@fjirsm.ac.cn}%
\thanks{$^{3}$College of Mechanical Engineering, Xinjiang University}%
\thanks{\textit{Proceedings of the 2022 IEEE 25th International Conference on Intelligent Transportation Systems (ITSC)}. Copyright by the author(s).}
}

\begin{document}

\maketitle
\thispagestyle{empty}
\pagestyle{empty}

\begin{abstract}

The annotation of 3D datasets is required for semantic-segmentation and object detection in scene understanding. In this paper we present a framework for the weakly supervision of a point clouds transformer that is used for 3D object detection. The aim is to decrease the required amount of supervision needed for training, as a result of the high cost of annotating a 3D datasets.  We propose an Unsupervised Voting Proposal Module, which learns randomly preset anchor points and uses voting network to select prepared anchor points of high quality. Then it distills information into student and teacher network. In terms of student network, we apply ResNet network to efficiently extract local characteristics. However, it also can lose much global information. To provide the input which incorporates the global and local information as the input of student networks, we adopt the self-attention  mechanism of transformer to extract global features, and the ResNet layers to extract region proposals. The teacher network supervises the classification and regression of the student network using the pre-trained model on ImageNet. On the challenging KITTI datasets, the experimental results have achieved the highest level of average precision compared with the most recent weakly supervised 3D object detectors.
\end{abstract}

\section{INTRODUCTION}

3D object detection has been a popular field with various applications, it functions as a major technology in scene understanding in automatic driving [16] . Due to the fact that 3D point clouds can reflect the in-depth information of 3D objects, researchers have taken it as starting the point to carry out research on 3D object detectors, which can express point clouds objects in the terms of normal 3D bounding box [17] . However, 3D object detection model training requires man ually labeling a large number of 3D bounding boxes around the irregular point clouds objects, which will lead to high cost of annotating thus to impede the application of 3D object detection. Currently, a large quantity of 3D object detectors [18] are mainly based on fully supervised learning and rely on a large number of 3D annotation truth boxes, which means limitations on scene applications that lack 3D labels. Therefore, study on weakly supervised or semi-supervised 3D object detectors with only a small number of 3D labels are practically significant. Researchers have also made certain achievements in their study on weakly supervised 2D object detectors and self-supervised 3D object detection [19] [20] . One of the superiorities of 3D detectors are detecting 3D object from point clouds in a weakly supervised way, and the 3D proposal bounding box, which is directly generated by the Unsupervised Proposal Module (UPM) of the VS3D [15] through the rotated and corrected preset anchor points. But one shortcoming of its performance is that it is not optimal due to the loss of geometric information from the virtual point clouds which is originated from the preset anchor points mapping. Moreover, the VGG16 network is not effective in serving the interests of teachers or students.
\\ ${\rm{\quad}}$In this work, we aim to solve these problems in VS3D, and the results of the experiment are better than those of VS3D. An overview of this paper follows a framework that we develop weakly supervised 3D object detection from point clouds transformer (WSPCT3D). First project 3D object proposals generated by the Unsupervised Voting Proposal Module (UVPM) onto the XYZ-map that relates to the point cloud in the same scene. Second, to blend global and local proposals, the backbone network of the student network leverages both the Self-Attention of the transformer and ResNet to merge the high-quality filtered proposal boxes with the XYZ-map generated after the input point clouds. Third, due to the lack of ground truth supervision, the precision of object recognition and classification using only student networks is not high, and the generalization ability among different datasets and settings is weak. To solve the problem, we apply a pre-trained teacher model to supervise the student network, with the student network modeling the behavior of the teacher during the training process. We also perform extensive experiments on the challenging KITTI [25] datasets to validate the proposed approach and its components. Promising results were based on several evaluation indicators. \textbf{Our method does not adopt any ground truth labels of 3D bounding boxes as the supervision of the training process}. Our experimental results show that is better than that of SOTA model VS3D.
\\${\rm{\quad}}$To summarize, our contributions are two-fold:
\begin{itemize}
\item[•]Through PointNet and widespread Hough voting, our UVPM can select preset anchors of the highest quality, which help to accurately predict downstream 3D bounding boxes.
\end{itemize}\begin{itemize}\item[•]The proposals from ResNet and Self-Attention fully capture the global and local features of the scene. The fusion of these proposals in the student network greatly improve the average precision of the 3D object detectors.
\end{itemize}

\section{RELATED WORK}
The purpose of 3D object detection is to detect the objects of interest and locate the 3D bounding boxes of the them. [1] proposes to extract 4D point clouds and detector boxes through the combination of 3D spatial and 1D temporal. PVGNet [2] voxelizes the point clouds to produce multi-level voxel features. [3] effectively combines features based on voxel and point to predict 3D bounding box. [4] leverages the disentangling transformation for 2D and 3D detection losses to produce high-quality 3D bounding boxes. [5] proposes a decoupling model of feature learning and external object prediction. [6] leverages the categorical depth distribution network to obtain the depth interval of the pixels. However, most of the existing methods are based on sufficient ground truth.
\\ ${\rm{\quad}}$Weakly supervised object detection assumes that the instance-level bounding boxes annotations are not provided by the training set. The weakly supervised object detection is realized using a small amount of labeled data to supervise a large amount of other unlabeled data. [7] alleviates the huge annotation burden and accurately predicts the object bounding box according to the point annotation. [8] describes the noisy label data with uncertainty, and alleviates the attention imbalance under data of different difficulty levels, thus can well fit all unmarked images. [9] proposes an efficient jointly thing-and-stuff mining (JTSM) framework for weakly supervised panoptic segmentation. [10] proposes a novel coupled multiple instance detection network(C-MIDN), which combines a pair of MIDNs with proposal removal. [11] tries to learn the texture 3D model of 3D reconstruction to solve the problem of 3D object detection, but its purpose is not to locate the 3D bounding box of the object. Learning 3D detection without complete supervision is more challenging, which will be explored in this paper.
\\ ${\rm{\quad}}$Knowledge distillation [12] is extensively used in transferring supervision cross modalities, for example [13] proposes a novel transferable semi-supervised 3D object detection network, which trains the backbone network to make class-agnostic segmentation and class-conditioned initial 3D box predictions on the strong classes in a fully-supervised manner. [14] proposes a weakly supervised pose estimation method using only 2D landmark annotation, which poses no restriction on the architecture of the student pose estimation network, and realizes accurate 3D pose estimation. [15] proposes a new Unbiased Mean Teacher (UMT) model for cross-domain object detection, which improves the effectiveness of knowledge distillation between teacher model and student model, and eliminate the impact of model bias. [16] proposes a cross-model data enhancement method to train the 3D object detector, and it uses the high-dimensional features of CNN to fuse with the point cloud of 3D detector. At present, cross-model learning is mainly applied to 2D recognition tasks, such as 2D detection and segmentation, while the 3D geometric information in depth data is not fully utilized. In addition, [34-37] recently proposed deep networks on point clouds for point clouds feature information learning. In the current research of weakly supervised 3D object detection, the interest of backbone network is not high, therefore our method explores how to characterize the object information in 3D more effectively.

\begin{figure*}[!htp]
\centering{\includegraphics[height=8.25cm,width=17.7cm]{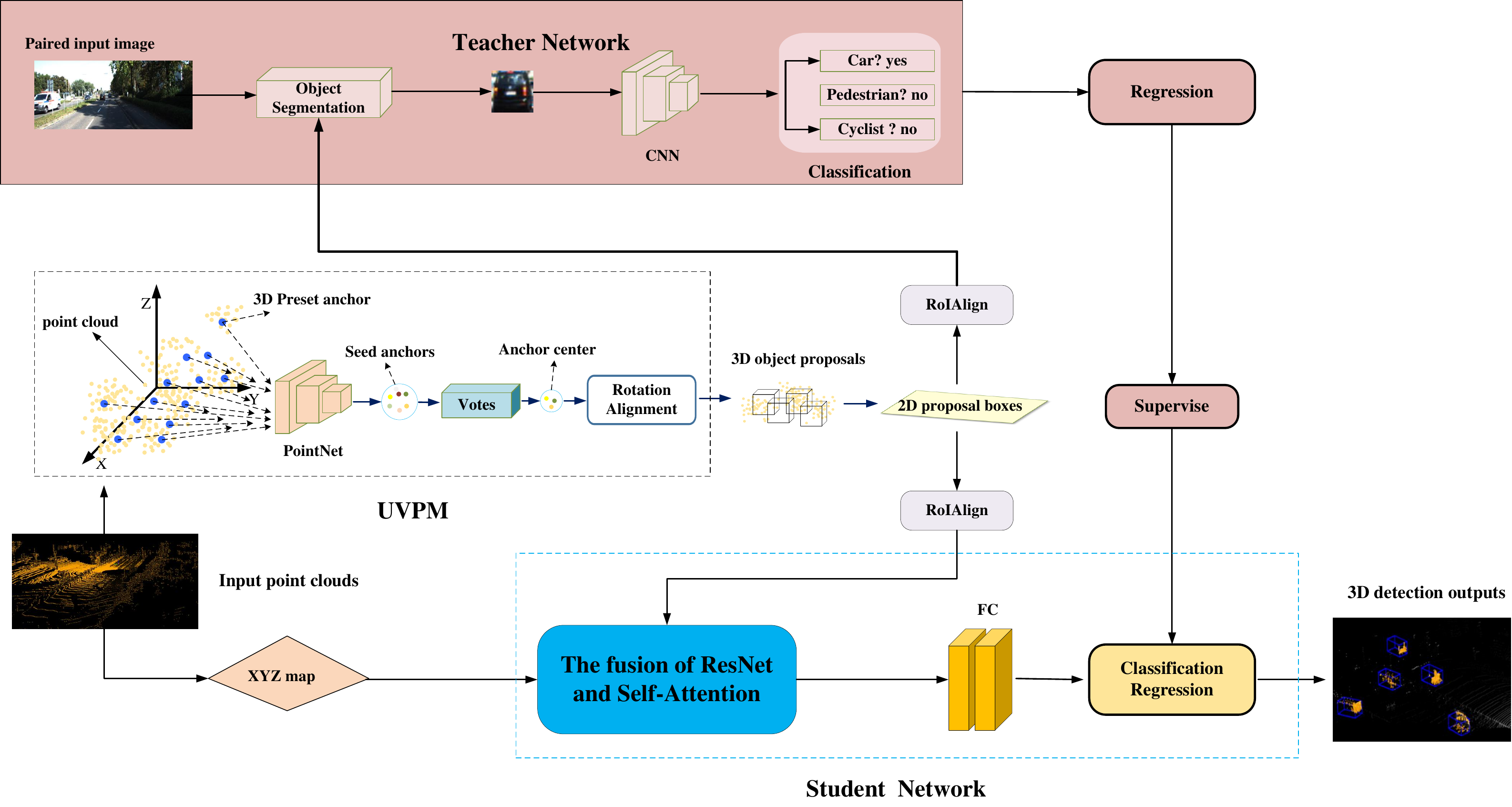}}
\caption{Overview of the Weakly Supervised Point Clouds Transformer for 3D Object Detection. There are three key points: firstly, in the teacher network, the input images are fused by CNN and the 2D proposal boxes generated by the UVPM module based on the normalized point clouds density [15] to generate 2D object bounding boxes. Secondly, the input point clouds are projected into a 2D point clouds XYZ map. The fusion module in the student network is fused with the 2D proposal boxes generated by the module from UVPM. Finally, the high-quality 3D object bounding boxes are obtained under the supervision of the teacher network. } \label{image}
\end{figure*}

\section{APPROACH}
\subsection{Overview}
The process of object detection can be defined as first extracting feature information from the input point clouds, and then it outputs the 3D bounding boxes of the object in the scene after model training. The existing 3D object detectors training process depends on data labels, while our weakly supervised 3D object detector does not need any labels in the training process. As shown in Figure 1, in the first stage, we propose an Unsupervised Voting Proposal Module (UVPM), and the 3D object proposals output by UVPM indicate that the objects are potentially contained in the regions from the input point clouds. In the second stage, the 3D object proposals are distilled to the student network and the teacher network respectively. In particular, in the student network, we introduce the specific details of the fusion of ResNet and Self-Attention, while the teacher network adopts the pre-trained model.
\subsection{Unsupervised Voting Proposal Module}
Lacking the geometry information of the anchors, Unsupervised Proposal Module (UPM) [15] can cause the failure to select the excellent ones for the 3D object proposals, generated by UPM random rotation,  which can lead to the loss of continuous characteristic of local proposals. Therefore, to solve the problem of the UPM, this paper proposes a UVPM, where 3D object proposals are output indicating that it is potentially contained within the regions from the input point clouds. The steps of how the UVPM works are as the following: Firstly, the preset anchors are placed at 0.2m intervals in the plane regions of [-35m, 35m]x[0, 70m] without the ground truth supervision, next the potential regions are filtered by the normalized point clouds density [15] threshold as object proposals regions, and then the selected anchors from the object proposals are mapped as pseudo point cloud of them. Secondly, the pseudo point clouds of the anchors passes the PointNet network to be output as a subset of  $M < N$ seed anchors information of the input anchor-pseudo point clouds, and then the local and global generated characteristics of each anchor in the object proposal regions can be extracted. The generated the seed anchors, with a significant level of dimensionality, they pass the Votes network so that those of the higher-quality can be selected. After that the subset of seed anchors are grouped into $K$ voting clusters, which can predict the central parameters of the bounding boxes of 3D objects proposals for each voting clusters group. Furthermore, we make the centers rotate and align to enhance the quality of the 3D object proposals. Finally, they are projected to become the 2D object proposals which are simultaneously distilled to the teacher network and the student network. \textbf{It is worth noting that where selection of seed anchors by voting in the 2nd step are the most essential difference between UVPM and UPM, where the UPM is to rotate and calibrate the center of the candidate boxes directly after the first step}.
\subsubsection{\textbf{PointNet and Voting}} The size of pseudo point clouds of the anchors of is N ×3, with a 3D coordinate for each of the N preset anchors, we aim to generate M voting anchors, where each preset anchor has both a 3D coordinate and a high dimensional feature vector. In the first stage, generating an accurate voting anchor requires geometric reasoning and contexts. Instead of relying on hand-crafted features, we adopt PointNet as our backbone due to its simplicity and it has demonstrated success on tasks ranging from normal estimation, semantic segmentation to 3D object localization [31-33]. The backbone network has several set-abstraction layers and feature propagation (upsampling) layers with skip connections, which outputs a subset of the preset anchors with XYZ and an enriched C-dimensional feature vector. The result is the generation of M seed anchors of dimension (3 + C). In the second stage, VoteNet [21] is a 3D point clouds single-stage object detector. To reduce the weight, VoteNet first processes the seed anchors $ \{ {x_i}\}_{i=1}^{N}  $ to generate a sub sampling set composed of $ M<N $ object loss combinations: the votes are divided into $K$ voting clusters, and each seed point votes on the center of the object to which it belongs. In the end, the 3D bounding box parameter $ {b}^ {(k)} $, the corresponding objectiveness score $s_k = P$, and the probability distribution ${p_{cls}}$ on $L$ potential semantic classes are predicted from each of the $K$ voting clusters. The bounding box parameter b is its center position $c\in R^3$, scale $d\in R^3$ and vertical axis direction $\Delta$. VoteNet applied Non-maximum suppression (NMS) to obtained data bounding boxes depending on the object score during the test. 
\subsubsection{\textbf{Anchor selection alignment}}
The normalized point clouds density $ D_c $ is not influenced by the distance because the front-view patch to the same size. When a target object contains an anchor, each $ D_c $ of the anchor should be above a certain threshold $ \delta $. However $ D_c<\delta $, determines the anchor does not contain a target object, the anchor is removed from object detection. To make a new anchor, we increase the remaining anchor by $ \emph{1}+\epsilon $ after they have been removed. The new anchor is accepted as a proposal if it contains a target object that are not present in the original preset anchor. The new anchor can verify that the anchor constraint is a wholly identified item, not just a portion of it. Among such, make sure the anchor is minor enough that the new anchor does not include the point clouds of the item that was identified.
\begin{figure*}[htp]
\centering{\includegraphics[height=8cm,width=17cm]{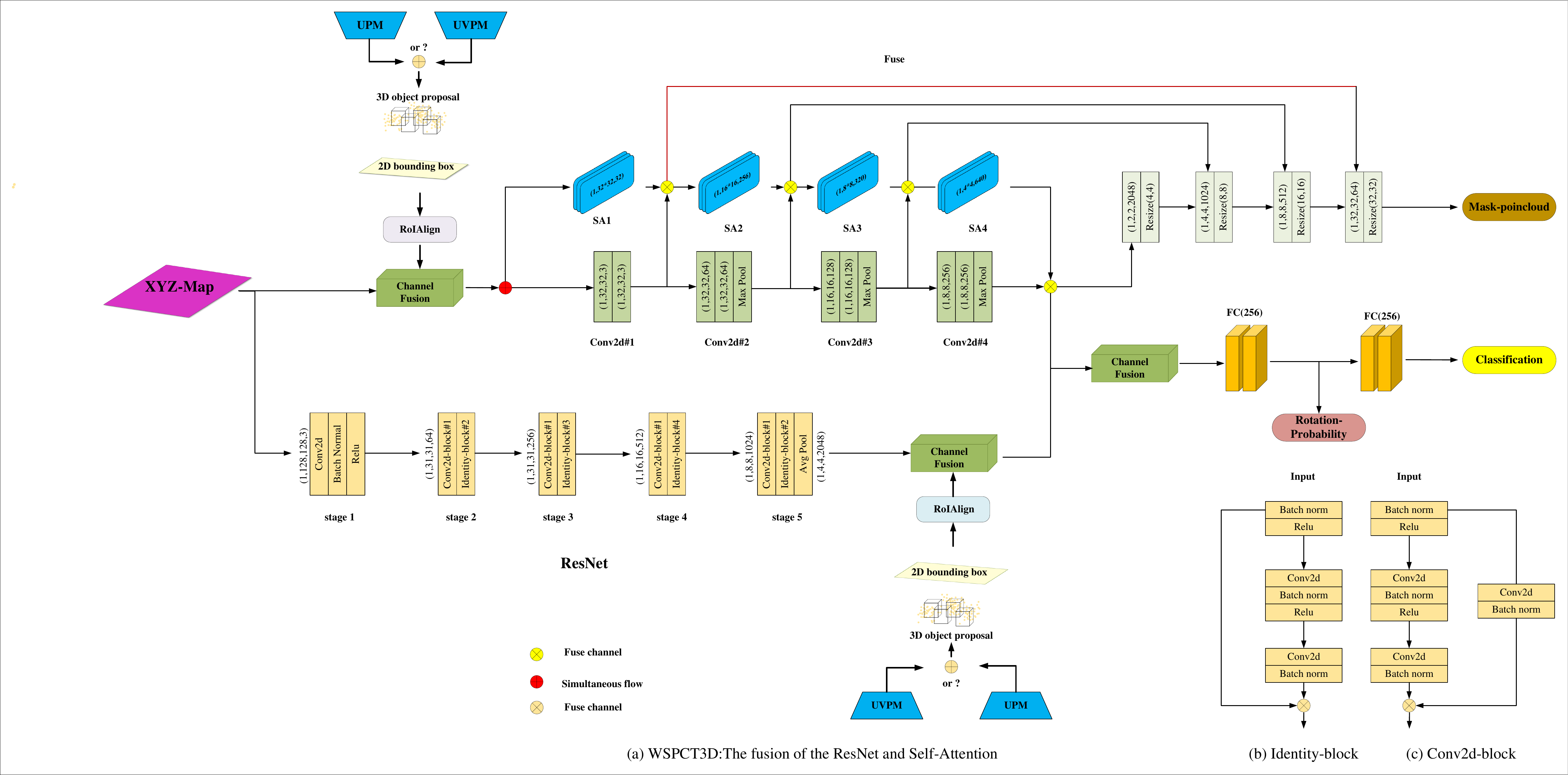}}
\caption{The fusion of ResNet and Self-Attention in the student network as shown in (a). The expanded structure of the Identity-block in ResNet is shown in (b), and the expanded structure of the Conv2d-block in ResNet is shown in (c). } \label{image}
\end{figure*}
\subsection{XYZ-map}
The front view representation complements the bird view representation by providing additional information. Because the LIDAR point clouds are sparse, projecting it onto the image plane produces a sparse 2D point map. As a result, before inputting the backbone, we convert the input point cloud into a dense front view XYZ map, which is similar to work [22]. In the front view map, the coordinate $ p_v = (r, c) $ of 3D point $ p = (x, y, z) $ can be represented as:

\begin{equation}   c = \lfloor a tan 2(y,x)/ \Delta\theta \rfloor  \end{equation}

\begin{equation}   r = \lfloor a tan 2(z, \sqrt{x^2+y^2} )/ \Delta \varphi \rfloor \end{equation}

Where $\Delta\theta$ and $\Delta \varphi$ are the horizontal and vertical resolution of laser beams, respectively. Three-channel features (height, distance and intensity) are used to encode the front view.
\begin{figure}[!htp]
\centering{\includegraphics[height=7cm,width=8cm]{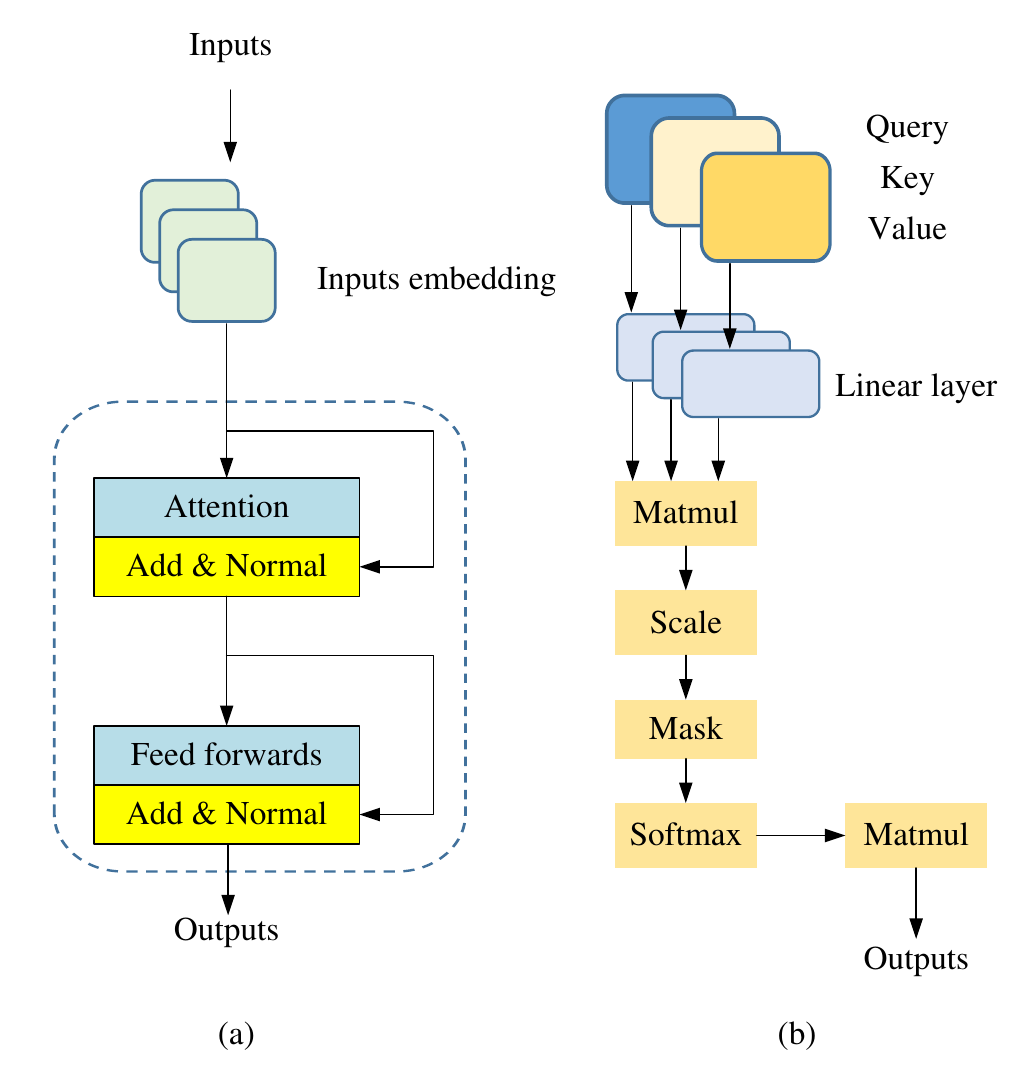}}
\caption{The structure of an encoder layer is shown in (a), the single-head attention mechanism of the encoder layer is shown in (b).} \label{image}
\end{figure}
\subsection{Student network}
The student network, as shown in Figure 2, mainly includes the ResNet and the Self-Attention based on transformer, the RoIAlign [23] layer, and the two fully connected layers in the second stage of 3D object detector. As the structure of ResNet is described in detail in Figure 2,  therefore this section focuses on the principles of the Self-Attention mechanism based on Transformer. The Self-Attention mechanism can be viewed of as mapping a query and a gathering of key value pairs to the output, with the query, key, value, and outcome all representing vectors as shown in Figure 3. The output is a weighted sum of values, with the weight being assigned to each value determined by the compatibility function of query and its corresponding key. Instead of applying the key, value, and query, it is advisable to apply various learning linear projections to linearly project the query, key, and value to the $d_k$, $d_k$ and $d_v$ dimensions, respectively. The attention function in parallel on each projected version of the query, key, and value to generate $d_v$ dimensional output values. To get the final outcome, these values are linked and projected again.
\begin{equation}  head_1=Attention(QW_i^Q,KW_i^K,VW_i^V)  \end{equation}
Where the projections are parameter matrices
\begin{equation}  MultiHead(Q,K,V)=Concat(head_1,\cdots,head_h)W^o  \end{equation}
Where:
$$ W_i^Q \in {\mathbb{R}}^{d_{\mathrm{mod} el\times  d_k }}$$
$$ W_i^K \in {\mathbb{R}}^{d_{\mathrm{mod} el\times  d_k }}$$
$$ W_i^V \in {\mathbb{R}}^{d_{\mathrm{mod} el\times  d_v }}$$
$$ W^O \in {\mathbb{R}}^{hd_v \times d_{\mathrm{mod} el }}$$
\begin{equation}   Attention(Q,K,V)=soft\mathrm{max(} \frac{QK^T}{\sqrt{d_k}})V    \end{equation}
\subsection{Teacher network}
\subsubsection{\textbf{Overview}}
The image-based teacher network uses VGG16 as the backbone for image classification and viewpoint regression tasks, with pre-training on ImageNet [30] and PASCALVOC [29] using the image level classification labels provided [24]. The teacher network accepts an image of no more than one item as input, next classifies it as background or a group of objects, and then returns the object viewpoint to its original rotation. A multi-box classification issue is referred to as viewpoint regression. The probability of 16 corner boxes split from a unit circle is predicted. The regression model of the angle of all boxes is the outcome of rotation. Teacher network is used to train a 3D object detection model as a prepared model.
\subsubsection{\textbf{from teacher to student}}
Each object proposes two predictions from teacher and student during the training process. Student imitate the confidence of the teacher network using the corrected cross entropy loss [15]. However, the teacher network may be ambiguous when the ability of the ready-made teacher network extracts the student network of various datasets. Firstly, if the teacher network is uncertain about its own output information, it information will not be passed on to student. Therefore, the classification score will not be applied to a training student branch if the classification score of the teacher network is in the ambiguous regions. Secondly, in the training process, we can extract recognition information from the teacher network and transfer it to the student network by matching the RGB image of the object with the point cloud. To be more specific, we project each generated object proposal by UVPM to an RGB image and a front view XYZ map. Furthermore, students use the corrected cross entropy loss to imitate the confidence of teachers network. Then we apply the teacher network to recognize the item proposal by removing the projection on the image. Simultaneously, we apply RoIAlign [23] to extract each proposed coding feature from the student backbone network and give these characteristics to the fully connected layers in order to predict the 3D bounding box of object.
\begin{table*}[!htp]
\begin{center}
\textbf{Table 1}\quad Compare the AP value of the five series methods of WSPCT3D under IoU threshold 0.3, and the five series methods of WSPCT3D are composed of UVPM, ResNet, SA and UPM [15] . In addition, SA2 means there are two Self-Attention blocks.\\
\setlength{\tabcolsep}{4mm}{
\begin{tabular}{c|llll}
\hline
\textbf{Model}                                                         & \textbf{INPUT}                & \multicolumn{1}{c}{\textbf{}}     & \multicolumn{1}{c}{\textit{\textbf{Recall(IOU=0.3)}}} & \multicolumn{1}{c}{\textbf{}}     \\ \hline
\textbf{}                                                              & \multicolumn{1}{c}{}          & \multicolumn{1}{c}{\textit{$AP_{2D}$}} & \multicolumn{1}{c}{\textit{$AP_{bird}$}}           & \multicolumn{1}{c}{\textit{$AP_{3D}$}} \\
& \multicolumn{1}{c}{\textbf{}}& Easy \; Moderate\; Hard    & Easy \; Moderate\; Hard     & Easy \; Moderate\; Hard             \\
\textbf{}   \\  
\texttt{}{VS3D{[}15{]}}                                                  & \texttt{}{Mono}                 & 77.73 \quad   73.82 \quad   65.71 & -  \qquad\quad  -     \quad\qquad  -  & 55.90 \quad 48.83 \quad 40.92 \\
\texttt{}{VS3D{[}15{]}}                                                  & \texttt{}{Stereo}               & 79.04 \quad   75.90 \quad   67.55 & -   \qquad\quad  -     \quad\qquad  -  & 70.72 \quad  63.78 \quad   52.03            \\
\texttt{}{VS3D{[}15{]}}                                                  & \texttt{}{Lidar}                & 78.64 \quad    74.41 \quad   66.24              & -  \quad\qquad      -  \quad\qquad         -                              & 65.96 \quad     59.76 \quad 49.78            \\
\textbf{\begin{tabular}[c]{@{}c@{}}Below are series of WSPCT3D\end{tabular}} & \multicolumn{1}{c}{\textbf{}} & \multicolumn{1}{c}{}              & \multicolumn{1}{c}{} & \multicolumn{1}{c}{}              \\
\texttt{}{UPM+ResNet53+SA2(1)}                                           & \texttt{}{Lidar}                & 81.75 \quad    74.78 \quad   66.68              & 70.42  \quad   63.52 \quad  53.55  & 69.01    \quad  61.78  \quad   51.72            \\
\texttt{}{UPM+ResNet50+SA2(2)}                                       & \texttt{}{Lidar}                & 82.08 \quad   77.25 \quad  68.64              & 72.03 \quad   65.04 \quad  55.65  & 70.81 \quad    63.21 \quad   53.85 \quad            \\
\texttt{}{UPM+ResNet53(3)}                                               & \texttt{}{Lidar} & 83.15 \quad   76.04 \quad  68.05              & 73.25 \quad   64.16 \quad  53.44   & 71.78 \quad    62.12 \quad   52.46            \\
\texttt{}{UPM+ResNet50+SA4(4)}                                           & \texttt{}{Lidar}                & 83.02 \quad   78.10 \quad  69.02              & 75.04   \quad66.35   \quad  56.94  & \textbf{75.34  \quad65.15  \quad55.51}   \\
\texttt{}{UVPM+ResNet50+SA4(5)}                                          & \texttt{}{Lidar}                & \textbf{84.21  \quad79.65  \quad 70.35}     & \textbf{76.21  \quad67.25  \quad56.32}                 & 74.04 \quad    64.27 \quad   54.70            \\ \hline
\end{tabular}}
\end{center}
\end{table*}

\begin{table*}[!htp]
\begin{center}
\textbf{Table 2}\quad Compare the AP value of the five series methods of WSPCT3D under IoU threshold 0.5, and the five series methods of WSPCT3D are composed of UVPM, ResNet, SA and UPM [15] . In addition, SA2 means there are two Self-Attention blocks.\\
\setlength{\tabcolsep}{4mm}{
\begin{tabular}{c|llll}
\hline
\textbf{Model}                                                         & \textbf{INPUT} & \multicolumn{1}{c}{\textbf{}}     & \multicolumn{1}{c}{\textit{\textbf{Recall(IOU=0.5)}}} & \multicolumn{1}{c}{\textbf{}}  \\ \hline
\textbf{}                                                              &                & \multicolumn{1}{c}{\textit{$AP_{2D}$}} & \multicolumn{1}{c}{\textit{$AP_{bird}$}}           & \multicolumn{1}{c}{$AP_{3D}$}       \\                                                           
\multicolumn{1}{l|}{\textbf{}}                                         &                & Easy \;  Moderate\;Hard              & Easy \;  Moderate \;Hard                           & Easy \;  Moderate \;Hard           \\
\textbf{} \\   
\texttt{}{VS3D{[}15{]}}                                                  & \texttt{}{Mono}  & 76.93 \quad    71.84 \quad    59.39             & - \quad \qquad      - \quad \qquad      -                              & 31.35 \quad    23.92 \quad    19.34          \\
\texttt{}{VS3D{[}15{]}}                                                  & \texttt{}{Stero} & 79.03  \quad  \textbf{72.71} \quad 59.77             & - \quad \qquad      - \quad \qquad      - & 40.98 \quad    34.09 \quad    27.65          \\
\texttt{}{VS3D{[}15{]}}                                                  & \texttt{}{Lidar} & 74.54 \quad    66.71 \quad    57.55             & - \quad \qquad      - \quad \qquad      - & 40.32 \quad    37.36 \quad    31.09          \\
\textbf{\begin{tabular}[c]{@{}c@{}}Below are series of WSPCT3D\end{tabular}} & \textbf{}      &                                   & \multicolumn{1}{c}{}                          & \multicolumn{1}{c}{}           \\
\texttt{}{UPM+ResNet53+SA2(1)}                                           & \texttt{}{Lidar} & 76.05 \quad    67.53 \quad    58.52             & 52.41 \quad   50.02 \quad   40.25                           & 41.21 \quad    40.03 \quad    33.15          \\
\texttt{}{UPM+ResNet50+SA2(2)}                                       & \texttt{}{Lidar} & 76.07 \quad    71.21 \quad    59.07             & 52.01 \quad   49.78 \quad   42.04                           & 43.12 \quad    41.34 \quad    32.16          \\
\texttt{}{UPM+ResNet53(3)}                                               & \texttt{}{Lidar} & 77.69 \quad    68.20  \quad   58.61 \quad              & 57.82 \quad   50.43 \quad   41.02                           & 48.41   \quad  40.22 \quad    33.20  \quad          \\
\texttt{}{UPM+ResNet50+SA4(4)}                                           & \texttt{}{Lidar} & 78.24 \quad    72.35 \quad    63.65             & 62.25 \quad   53.52 \quad   45.41                           & \textbf{52.82  \quad43.10  \quad36.12 \quad } \\
\texttt{}{UVPM+ResNet50+SA4(5)}                                          & \texttt{}{Lidar} &\textbf{80.15} \quad72.66  \quad\textbf{64.98}            & \textbf{64.27} \quad\textbf{53.46} \quad\textbf{46.98}                           & 51.24  \quad  44.35  \quad  35.32   \\ \hline
\end{tabular}}
\end{center}
\end{table*}

\begin{table*}[!htp]
\begin{center}
\textbf{Table 3}\quad Compare the AP value of the previous methods under IoU threshold 0.3 and 0.5, and the two series methods of WSPCT3D are composed of UVPM, ResNet, SA and UPM [15] . In addition, SA4 means there are four Self-Attention blocks.\\
\setlength{\tabcolsep}{6mm}{
\begin{tabular}{c|llll}
\hline
\textbf{Model}             & \textbf{INPUT}  & \textit{\textbf{Recall(IOU=0.3)}}     & \textit{\textbf{Recall(IOU=0.5)}}     \\ \hline
\textbf{}                  &                 & \textit{$AP_{2D}$}                 & \textit{$AP_{2D}$}                 \\
\multicolumn{1}{l|}{\textbf{}}                                         &                & Easy \quad Moderate \,\quad Hard         & Easy \quad Moderate \,\quad Hard \\                         \\
\texttt{}{PCL{[}26{]}} & \texttt{}{Mono} &5.916\qquad 4.678\qquad  3.765         &1.878\qquad 1.058\qquad  0.935          \\
\texttt{}{OICR{[}27{]}}      & \texttt{}{Mono}   & 13.50\qquad  8.604\qquad   8.045\qquad          & 6.481\qquad  2.933\qquad   3.270          \\
\texttt{}{MELM{[}28{]}}      & \texttt{}{Mono}   & 8.054\qquad  7.282\qquad   6.882\qquad          & 2.796\qquad  1.486\qquad   1.476\qquad          \\
\texttt{}{VS3D{[}15{]}}      & \texttt{}{Mono}   & 77.73\qquad  73.82\qquad   65.71          & 76.93\qquad  71.84\qquad   59.39          \\
\texttt{}{VS3D{[}15{]}}      & \texttt{}{Stereo} & 79.04\qquad  75.90\qquad   67.55          & 79.03\qquad  \textbf{72.71}\qquad   59.77          \\
\texttt{}{VS3D{[}15{]}}      & \texttt{}{Lidar}  & 78.64\qquad  74.41\qquad   66.24          & 74.54\qquad  66.71\qquad   57.55          \\
\textbf{Ours WSPCT3D}           & \textbf{}       &                               &                               \\
\texttt{}{UPM+ResNet50+SA4(4)}  & \texttt{}{Lidar}  &83.02\qquad  78.10\qquad    69.02          &78.24\qquad  72.35\qquad    63.65          \\
\texttt{}{UVPM+ResNet50+SA4(5)} &\texttt{}{Lidar} &\textbf{84.21}\qquad\textbf{79.65\quad}\quad\textbf{70.35}&\textbf{80.15}\qquad 72.66\qquad\textbf{64.98} \\ \hline
\end{tabular}}
\end{center}
\end{table*}

\section{EXPERIMENTS}

In this section, we first introduce some setup details about our experiment, including datasets, evaluation and implementation details. We will then go through the details and analysis of UVPM, ResNet and Self-Attention fusion details, and finally give high quality results. The experiments were conducted on KITTI datasets, which contain real image data from urban, rural and highway scenes, with up to 15 vehicles and 30 pedestrians in each image, along with various levels of occlusion. KITTI datasets provides multiple unified evaluation matrices, including average Precision (AP) and Recall rate curve at different intersection of Union (IOU) Thresholds and different occlusion level. Our experiment is based on the Linux, Tensorflow-gpu=1.12.0, Python 3.6 and TeslaV100S-PCIE platforms.
\subsection{Experimental environment and data set settings }
\subsubsection{\textbf{Details of UVPM experiment}}
The preset anchors are placed at 0.2m intervals in the plane area [-35m,35m]x[0,70m] without ground truth supervision. We use the region filtered by the normalized point clouds density threshold as object proposals region, which represent the anchor points in the object proposals region as pseudo point clouds. In the PointNet network, the pseudo point clouds pass through four point feature extraction layers in turn, then it generates a point cloud with additional features that is called the Seed Anchor after passing two Feature propagation(FP) Modules. The pseudo point clouds feature channels are set as 3-coordinate channels (X, Y, Z) and 58 additional channels, which are used as inputs of the first feature extraction layer together. Region is set as Raidus [0.2, 0.4, 0.8, 1.2] in turn. All multi-layer Perception(MLP) have the sizes of [64, 64, 128]. The sizes of the MLPS of the two FP layers are [128, 128] and [256, 256] respectively. The Seed Anchors are generated a small number of proposal anchors with additional features through Voting Module layers output as (2048, 4). It is suggested that anchors can be used as central candidate points of 2D Proposal bounding boxes, then passing the center rotation, alignment, and correction are carried out for the 2D Proposal bounding boxes, and we built the loss functions of rotation, alignment, and correction for evaluation.
\subsubsection{\textbf{The fusion of the ResNet and Attention}}
The fusion of the ResNet and Attention in the student network is shown in Figure 2. We begin cropping the xyz-map to the shape (1, 128, 128, 3) and input it into two branches: (1) First, 3D bounding boxes are generated by UPM or UVPM modules, and then features extracted from 2D bounding boxes by RoIAlign are integrated with XYZ-Map by adding feature channels while keeping the same batch length and width. We then input the fused features into the Self-Attention and Conv2D branches again by adding the feature channels. In the self-attention block, the dropout was 0.5, and the Conv1D convolution kernel was 4 in size and an Attention head. However, the number of Self-Attention (SA) blocks as shown in Figure 2(a) was 4. In addition, we set two Self-Attention blocks for comparison. The size of all convolution kernels in Conv2D blocks is [3, 3] , dilation-rate set as 1. It is worth noting that the number of SA blocks and Conv2D blocks changes in the same way. After that, the features fused with SA block and Conv2D block were simultaneously fused with the downstream up-sampled convolution layer channel, as shown in Figure 2(a). At last Mask-point clouds was output. (2) XYZ-Map was output after five stages of ResNet as the shape (1, 4, 4, 2048). And then the features extracted from 2D bounding box by RoIAlign are integrated in the way of feature channel addition. In ResNet networks, we set ResNet50 and ResNet53 for comparison. The only difference between these two networks is the number of convolutional layers. The size of all convolutional kernels is [3, 3], dilation-rate is 1. Finally, we fuse the features from SA block and Conv2D block in the first branch with those from the second branch. The probability of Rotation is output through two full connection layers, and Classification is output by two full connection layers. During the experiment, according to the number of convolution layers of UPM or UVPM, ResNet network and the fusion number of SA block and Conv2D block, we set up five comparison methods based on WSPCT3D, as shown in Table 1 and Table 2.
\begin{figure*}[!htp]
\centering{\includegraphics[height=7.8cm,width=16cm]{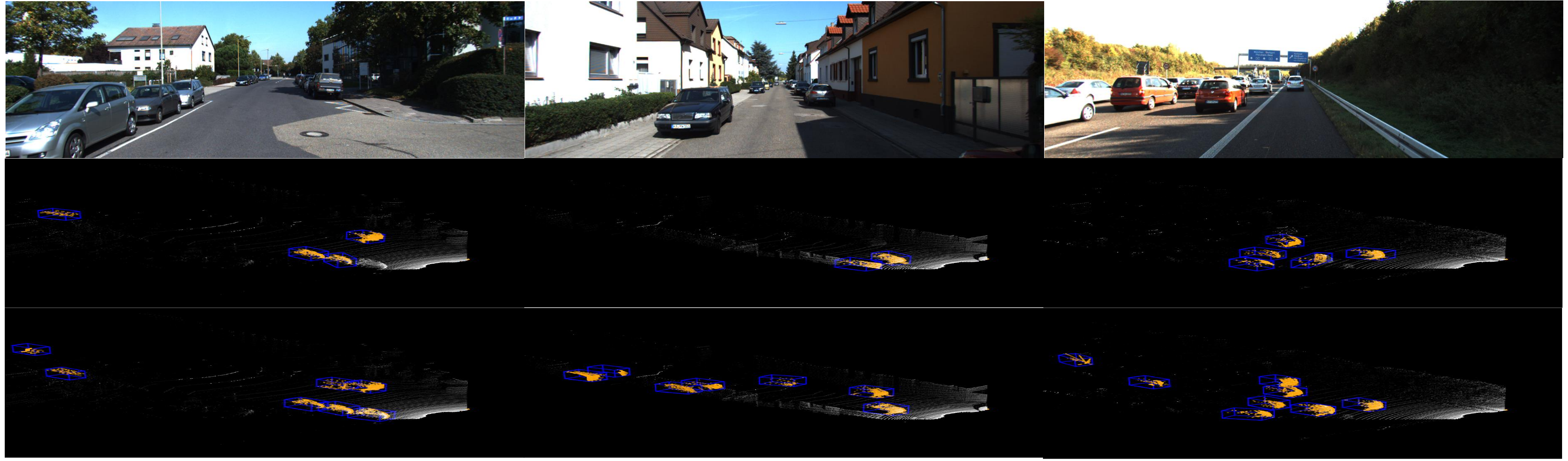}}
\caption{There are three different scenes from left to right, in each scene the prediction 3D bounding boxes in the middle are the Lidar version of the VS3D, and the prediction 3D bounding boxes at the bottom are our WSPCT3D: UPM+ResNet50+SA4(4) method. } \label{image}
\end{figure*}
\subsection{ Qualitative Analysis and Result}
Our WSPCT3D series method is carried out where the type of input signal is Lidar scans while VS3D, which is set as the baseline for comparison, includes three types of input signals including Lidar scans, Stereo images, and Monocular images.
\\${\rm{\quad}}$In Table 1, under IoU threshold 0.3, first of all, the AP value of our method (1), (2) and (3) is comprehensively superior to that of the Lidar version of VS3D, indicating that the method of fusing ResNet with SA block has achieved good results. Specifically, methods (1), (2) and (3) are improved to the range of [0.37, 4.51], and [1.94, 5.82] compared with the that of Lidar version of VS3D under $AP_{3D}$, which means that our method outperforms VS3D by 1.94 to 5.82 under current standards. In addition, the comparison among methods (1), (2) and (3) shows that there is no significant the effect of only two SA blocks on the experimental results under $AP_{3D}$ and the fusion of ResNet50 or ResNet53 makes no significant improvement on the experimental effect neither. We believe that the global feature information that SA block processes is incomplete or not enough, therefore, as shown in Method (4) in term of the number of SA blocks, our method outperforms the Lidar version of VS3D by 5.39 at least and 9.38 at most. Our UVPM module, as shown in method (5), is improved by the range of [4.51, 8.05] compared with the Lidar version of VS3D achieves the best effect under $AP_{2D}$. However, the comparison between methods (4) and (5) shows that the performance of UVPM under Lidar $AP_{3D}$ is slightly inferior to that of UPM, so our UVPM needs to be improved. Second, in Table 1, the Stereo version of VS3D all exceed our method (1). However, our WSPCT3D series methods (2), (3), (4) and (5) have been improved compared with the Stereo version of VS3D, although the extent of improvement is not very high compared with the Lidar version of VS3D under $AP_{3D}$. It is also shown in Table 1 that the type of input signal has an impact on the evaluation of the overall model. For example, the signal of Lidar input has a richer representation of 3D target information than Stereo, so it is better than Stereo under $AP_{3D}$. Whereas, in VS3D series methods, Stereo has a better effect than Lidar under $AP_{2D}$ and $AP_{3D}$. We analyzed that this situation was caused by a low IOU threshold of 0.3.\\
${\rm{\quad}}$In Table 2, the IOU threshold 0.5. Firstly, the AP values of our method (1), (2) and (3) are better than the Lidar version of VS3D. Specifically, methods (1), (2) and (3) are improved by the range of [0.97, 3.15] compared with the Lidar version of VS3D under $AP_{2D}$ on the same basis, and [1.94,  5.82] compared with that under $AP_{3D}$. It can be seen that compared with the Lidar version of VS3D that under $AP_{2D}$, our method (4) has been improved to the range of [2.78, 4.38]. In particular, compared with the Lidar version of VS3D under $AP_{3D}$, it has been improved by the range of [5.04, 12.5] to achieve the best effect under $AP_{3D}$. In addition, our UVPM module in method (5) under $AP_{2D}$ is improved by the range of [5.61, 7.43] compared with the Lidar of VS3D under $AP_{2D}$, which achieves the best results under $AP_{2D}$. However, the comparison between (4) and (5) shows that the performance of UVPM under $AP_{3D}$ is still slightly inferior to that of UPM, and our UVPM still needs to be improved. However, the Stereo version of VS3D under $AP_{2D}$ in Table 2 exceeds our methods (1), (2) and (3), and methods (4) and (5) are improved by the range of [-0.36, 5.21] compared with it. In addition, Table 3 presents that our method (4) and (5) have advantages over PCL, OICR, MELM and VS3D, but it also shows that our method has less improvement than the Stereo version of VS3D under $AP_{2D}$ on the same basis. According to our analysis, the relative absence of evaluation information was caused by inconsistent input signal types. Considering that our WSPCT3D series of methods are all evaluated as Lidar input, with Lidar scans being the signal some information was lost in the process of projecting 3D Lidar points cloud into 2D point clouds. As for Stereo images input signal, 3D point clouds are generated by 2D images through calculation of values of different perspectives and then projected as 2D point clouds. Therefore, $AP_{2D}$ calculation itself has more advantages over 2D Lidar, so we are confident that all our WSPCT3D series methods have better performance in calculating $AP_{2D}$ when the input is signal types images. Different from Table 1, compared with the Stereo version of VS3D under $AP_{3D}$, our methods (1), (2) and (3) are comprehensively improved by the range of [0.23,7.43] in $AP_{2D}$ evaluation. It is worth noting that method (4) is improved by the range of [8.47,11.82] compared with the Stereo version of VS3D under $AP_{3D}$, which indicates that our methods have great advantages especially in 3D object detection. Furthermore, as shown in Figure 4, through the prediction of the 3D bounding box of targets, we can observe that the effect of method (4) is significantly higher than that of the Lidar version of VS3D.

\section{CONCLUSION}
This paper studies the most recent 3D Weakly Supervised Object Detection algorithm and VS3D, after that a UVPM module is proposed  to cover information from the point clouds itself when the input is the anchor pseudo point clouds, thus to reduce the frequency of rotations and corrections in the UPM [15] and to improve the  accuracy and efficiency of the candidate boxes. The key point is that for the final classification output, we merge the ResNet local proposals with the global proposals from transformer Self-Attention. On The KITTI datasets, our WSPCT3D series method achieves the best results compared with existing weakly supervised methods. However, our method is limited to optimizing the student network, and the UVPM needs to be further improved. Therefore, in the future, we will include an instructor for the teacher network, and the Stereo version of the WSPCT3D.


\begin{thebibliography}{00}
\bibitem{b1} Charles R. Qi, Yin Zhou, Mahyar Najibi, Pei Sun, Khoa Vo, Boyang Deng, and Dragomir Anguelov. ``Offboard 3d object detection from point cloud sequences'', in\emph{ IEEE conference on computer vision and pattern recognition(CVPR)}. pp. 6134–6144, 2021.
\bibitem{b2} Zhenwei Miao, Jikai Chen, Hongyu Pan, Ruiwen Zhang, Kaixuan Liu, Peihan Hao, Jun Zhu, Yang Wang, and Xin Zhan. `` Pvgnet: A bottom-up one-stage 3d object detector with integrated multi-level features'', in\emph{ IEEE conference on computer vision and pattern recognition (CVPR)}. pp. 3279–3288, 2021.
\bibitem{b3} Zetong Yang, Yanan Sun, Shu Liu, Xiaoyong Shen, and Jiaya Jia. ``Std: sparse-to-dense 3d object detector for point cloud'', in\emph{ IEEE conference on computer vision and pattern recognition (CVPR)}. pp. 1951–1960, 2019.
\bibitem{b4} Andrea Simonelli, Samuel Rota Bulo, Lorenzo Porzi, Manuel Lopez-Antequera, and Peter Kontschieder. ``Disentangling monocular 3d object detection'', in\emph{ IEEE conference on computer vision and pattern recognition (CVPR)}. pp. 1991–1999, 2018.
\bibitem{b5} Yunpeng Zhang, Jiwen Lu, and Jie Zhou. ``Objects are different: flexible monocular 3d object detection'', in\emph{ IEEE conference on computer vision and pattern recognition (CVPR)}. pp. 3289–3298, 2021.
\bibitem{b6} Jiyang Gao, Jiang Wang, Shengyang Dai, Li-Jia Li, and Ram Nevatia. `` Note-rcnn: Noise tolerant ensemble rcnn for semi-supervised object detection'', in\emph{ IEEE conference on computer vision and pattern recognition (CVPR)}. pp. 9508–9517, 2018.
\bibitem{b7} Zhenyu Wang, Yali Li, Ye Guo, Lu Fang, and Shengjin Wang. ``Data-uncertainty guided multi-phase learning for semi-supervised object detection'', in\emph{ IEEE conference on computer vision and pattern recognition (CVPR)}. pp. 4568–4577, 2021.
\bibitem{b8} Yan Gao, Boxiao Liu, Nan Guo, Xiaochun Ye, Fang Wan, Haihang You, and Dongrui Fan. ``C-midn: coupled multiple instance detection network with segmentation guidance for weakly supervised object detection'', in\emph{ IEEE/CVF International Conference on Computer Vision (ICCV)}. pp. 9834–9843, 2019.
\bibitem{b9} Philipp Henzler, Jeremy Reizenstein, Patrick Labatut, Roman Shapovalov, Tobias Ritschel, Andrea Vedaldi, and David Novotny. ``Unsupervised learning of 3d object categories from videos in the wild'', in\emph{ IEEE conference on computer vision and pattern recognition (CVPR)}. pp. 4700–4709, 2021.
\bibitem{b10} Nuno Garcia, Pietro Morerio, and Vittorio Murino. ``Modality distillation with multiple stream networks for action recognition'', in\emph{ Proceedings of the European Conference on Computer Vision (ECCV)}. pp. 106–121, 2018.
\bibitem{b11} Abrar H. Abdulnabi, Bing Shuai, Zhen Zuo, Lap-Pui Chau, and Gang Wang. ``Multimodal recurrent neural networks with information transfer layers for indoor scene labeling'', in\emph{ IEEE Transactions on Multimedia}. pp. 1656-1671, 2018.
\bibitem{b12} Yew Siang Tang and Gim Hee Lee. 2019. Transferable semi-supervised 3d object detection from rgb-d data. In\emph{ IEEE/CVF International Conference on Computer Vision (ICCV)}. 1931–1940.
\bibitem{b13} Jinhong Deng, Wen Li1, Yuhua Chen, and Lixin Duan. ``Unbiased mean teacher for cross-domain object detection'', in\emph{ IEEE conference on computer vision and pattern recognition (CVPR)}. pp. 4091–4101, 2021.
\bibitem{b14} Chunwei Wang, Chao Ma, Ming Zhu, and Xiaokang Yang. ``Point augmenting: cross-modal augmentation for 3d object detection'', in\emph{ IEEE conference on computer vision and pattern recognition (CVPR)}. pp. 11794–11803, 2021.
\bibitem{b15} Zengyi Qin, Jinglu Wang, and Yan Lu. `` Weakly supervised 3d object detection from point clouds'', in\emph{ ACM MM}. pp. 4144–4152, 2020.
\bibitem{b16} Zetong Yang, Yanan Sun, Shu Liu, Xiaoyong Shen, and Jiaya Jia. ``Std: sparse-to-dense 3d object detector for point cloud'', in\emph{ IEEE/CVF International Conference on Computer Vision (ICCV)}. pp. 1951–960, 2019.
\bibitem{b17} Yunsong Zhou, Yuan He, Hongzi Zhu, Cheng Wang, Hongyang Li, Qinhong Jiang. ``Monocular 3d object detection: an extrinsic parameter free approach'', in\emph{ IEEE conference on computer vision and pattern recognition (CVPR)}. pp. 7556–7566, 2021.
\bibitem{b18} Pei Sun, Weiyue Wang, Yuning Chai, Gamaleldin Elsayed, Alex Bewley, Xiao Zhang, Cristian Sminchisescu, and Dragomir Anguelov. ``Rsn: range sparse net for efficient, accurate lidar 3d object detection'', in\emph{ IEEE conference on computer vision and pattern recognition (CVPR)}. pp. 5725–5734, 2021.
\bibitem{b19} Pinchun Chen, Bohan Kung, Juncheng Chen. ``Class-aware robust adversarial training for object detection'', in\emph{ IEEE conference on computer vision and pattern recognition (CVPR)}. pp. 10420–10429, 2021.
\bibitem{b20} Jihan Yang, Shaoshuai Shi, Zhe Wang, Hongsheng Li, and Xiaojuan Qi. ``St3d: self-training for unsupervised domain adaptation on 3d object detection'', in\emph{ IEEE conference on computer vision and pattern recognition (CVPR)}. pp.  10368–10378, 2021.
\bibitem{b21} Charles R Qi, Or Litany, Kaiming He, and Leonidas JGuibas. ``Deep hough voting for 3d object detection in point clouds'', in\emph{ IEEE/CVF International Conference on Computer Vision (ICCV)}. pp. 9277–9286, 2019.
\bibitem{b22} Bo Li, Tianlei Zhang, Tian Xia. ``Vehicle detection from 3d lidar using fully convolutional network'',in\emph{ Robotics: Science and Systems(RSS)}. 2016.
\bibitem{b23} Kaiming He, Georgia Gkioxari, Piotr Dollar, and Ross Gir-shick. ``Mask r-cnn'', in\emph{ IEEE/CVF International Conference on Computer Vision (ICCV)}. pp. 2961–2969, 2017. 
\bibitem{b24} Yu Xiang, Roozbeh Mottaghi, and Silvio Savarese. ``Beyond pascal: a benchmark for 3d object detection in the wild'', in\emph{ IEEE Winter Conference on Applications of Computer Vision (WACV)}. pp. 75-82, 2014.
\bibitem{b25} Andreas Geiger, Philip Lenz, and Raquel Urtasun. ``Are we ready for autonomous driving? the kitti vision benchmark suite'', in\emph{ IEEE conference on computer vision and pattern recognition (CVPR)}. pp. 3354–3361, 2012.
\bibitem{b26} Peng Tang, Xinggang Wang, Song Bai, Wei Shen, Xiang Bai. `` Pcl: proposal cluster learning for weakly supervised object detection'', in\emph{ IEEE Transactions on Pattern Analysis and Machine Intelligence}. pp. 176–191, 2020.
\bibitem{b27} Peng Tang, Xinggang Wang, Xiang Bai, Wenyu Liu. `` Multiple instance detection network with online instance classifier refinement'', in\emph{ IEEE conference on computer vision and pattern recognition (CVPR)}. pp. 2843–2851, 2017.
\bibitem{b28} Fang Wan, Pengxu Wei, Jianbin Jiao, Zhenjun Han, and Qixiang Ye. ``Minentropy latent model for weakly supervised object detection'', in\emph{ IEEE conference on computer vision and pattern recognition (CVPR)}. pp. 1297–1306, 2018.
\bibitem{b29} Mark Everingham, Luc Gool, Christopher K. Williams, John Winn, and Andrew Zisserman. ``The Pascal Visual Object Classes (VOC) Challenge'', in \emph{International Journal of Computer Vision (IJCV)}. pp. 303–338, 2010.
\bibitem{b30} Olga Russakovsky, Jia Deng, Hao Su, Jonathan Krause, Sanjeev Satheesh, Sean Ma,Zhiheng Huang, Andrej Karpathy, Aditya Khosla, Michael Bernstein, Alexander C.Berg, and Li Fei-Fei. ``Imagenet large scale visual recognition challenge'', in \emph{International Journal of Computer Vision (IJCV)}. pp. 211–252, 2015.
\bibitem{b31} Paul Guerrero, Yanir Kleiman, Maks Ovsjanikov, and Niloy J Mitra. ``Pcpnet learning local shape properties from raw point clouds'', in \emph{Computer Graphics Forum}. pp. 75–85, 2018.
\bibitem{b32} Loic Landrieu and Martin Simonovsky. ``Large-scale point cloud semantic segmentation with superpoint graphs'', in \emph{Proceedings of the IEEE Conference on Computer Vision and Pattern Recognition(CVPR)}. pp. 4558–4567, 2018.
\bibitem{b33} Charles R Qi, Wei Liu, Chenxia Wu, Hao Su, and Leonidas Guibas. ``Frustum pointnets for 3d object detection from rgbd data'', in \emph{Proceedings of the IEEE Conference on Computer Vision and Pattern Recognition(CVPR)}. pp. 918–927, 2018. 
\bibitem{b34} Benjamin Graham, Martin Engelcke, and Laurens van derMaaten. ``3D semantic segmentation with submanifold sparse convolutional networks'', in \emph{Proceedings of the IEEE Conference on Computer Vision and Pattern Recognition(CVPR)}. pp. 9224–9232, 2018.
\bibitem{b35} Hang Su, Varun Jampani, Deqing Sun, Subhransu Maji, Evangelos Kalogerakis, Ming-Hsuan Yang, and Jan Kautz. ``Splatnet: Sparse lattice networks for point cloud processing'', in \emph{Proceedings of the IEEE Conference on Computer Vision and Pattern Recognition(CVPR)}. pp. 2530–2539, 2018.
\bibitem{b36} Yangyan Li, Rui Bu, Mingchao Sun, Wei Wu, Xinhan Di, and Baoquan Chen. ``Pointcnn: Convolution on x-transformed points'', in \emph{Neural Information Processing Systems}. pp. 828–838, 2018.
\bibitem{b37} Charles R Qi, Li Yi, Hao Su, and Leonidas J Guibas. ``Pointnet++:Deep hierarchical feature learning on point sets in a metric space''. arXiv preprint arXiv:1706.02413. 2017



\end{thebibliography}
\end{document}